\documentclass[twocolumn, switch]{article} 

\usepackage{preprint}

\usepackage{amsmath, amsthm, amssymb, amsfonts}

\usepackage[numbers,square]{natbib}
\usepackage[utf8]{inputenc}	
\usepackage[T1]{fontenc}	
\usepackage{xcolor}		
\usepackage[colorlinks = true,
            linkcolor = purple,
            urlcolor  = blue,
            citecolor = cyan,
            anchorcolor = black]{hyperref}	
\usepackage{booktabs} 		
\usepackage{nicefrac}		
\usepackage{microtype}		
\usepackage{lineno}		
\usepackage{float}			

\usepackage{lipsum}		

\usepackage{newfloat}
\DeclareFloatingEnvironment[name={Supplementary Figure}]{suppfigure}
\usepackage{sidecap}
\sidecaptionvpos{figure}{c}

\usepackage{titlesec}
\titlespacing\section{0pt}{12pt plus 3pt minus 3pt}{1pt plus 1pt minus 1pt}
\titlespacing\subsection{0pt}{10pt plus 3pt minus 3pt}{1pt plus 1pt minus 1pt}
\titlespacing\subsubsection{0pt}{8pt plus 3pt minus 3pt}{1pt plus 1pt minus 1pt}

\usepackage{tikz,xcolor,hyperref}

\definecolor{lime}{HTML}{A6CE39}
\DeclareRobustCommand{\orcidicon}{
	\begin{tikzpicture}
	\draw[lime, fill=lime] (0,0) 
	circle [radius=0.16] 
	node[white] {{\fontfamily{qag}\selectfont \tiny ID}};
	\draw[white, fill=white] (-0.0625,0.095) 
	circle [radius=0.007];
	\end{tikzpicture}
	\hspace{-2mm}
}
\foreach \x in {A, ..., Z}{\expandafter\xdef\csname orcid\x\endcsname{\noexpand\href{https://orcid.org/\csname orcidauthor\x\endcsname}
			{\noexpand\orcidicon}}
}

\title{Can Machines Read Coding Manuals Yet? -- A Benchmark for Building Better Language Models for Code Understanding}

\usepackage{authblk}

\author[1]{Ibrahim Abdelaziz}
\author[1]{Julian Dolby}
\author[2]{Jamie McCusker}
\author[1]{Kavitha Srinivas}

\affil[1]{IBM Research, T.J. Watson Research Center, Yorktown Heights, NY, USA\\

\{ibrahim.abdelaziz1, kavitha.srinivas\}@ibm.com, dolby@us.ibm.com}
\affil[2]{Rensselaer Polytechnic Institute (RPI), Troy, NY, USA

mccusj2@rpi.edu
}

\begin{document}

\twocolumn[ 
  \begin{@twocolumnfalse} 
  
\maketitle

\begin{abstract}
    Code understanding is an increasingly important application of Artificial Intelligence. A fundamental aspect of understanding code is understanding text about code, e.g., documentation  and forum discussions.  Pre-trained language models (e.g., BERT) are a popular approach for various NLP tasks, and there are now a variety of benchmarks, such as GLUE, to help improve the development of such models for natural language understanding. However, little is known about how well such models work on textual artifacts about code, and we are unaware of any systematic set of downstream tasks for such an evaluation.  In this paper, we derive a set of benchmarks (BLANCA - Benchmarks for LANguage models on Coding Artifacts) that assess code understanding based on tasks such as predicting the best answer to a question in a forum post, finding related forum posts, or predicting classes related in a hierarchy from class documentation.  We evaluate the performance of current state-of-the-art language models on these tasks and show that there is a significant improvement on each task from fine tuning.  We also show that multi-task training over BLANCA tasks helps build better language models for code understanding.
\end{abstract}
\vspace{0.35cm}

  \end{@twocolumnfalse} 
] 



\section{Introduction}
\begin{figure}
  \begin{center}
  \includegraphics[width=.85\linewidth]{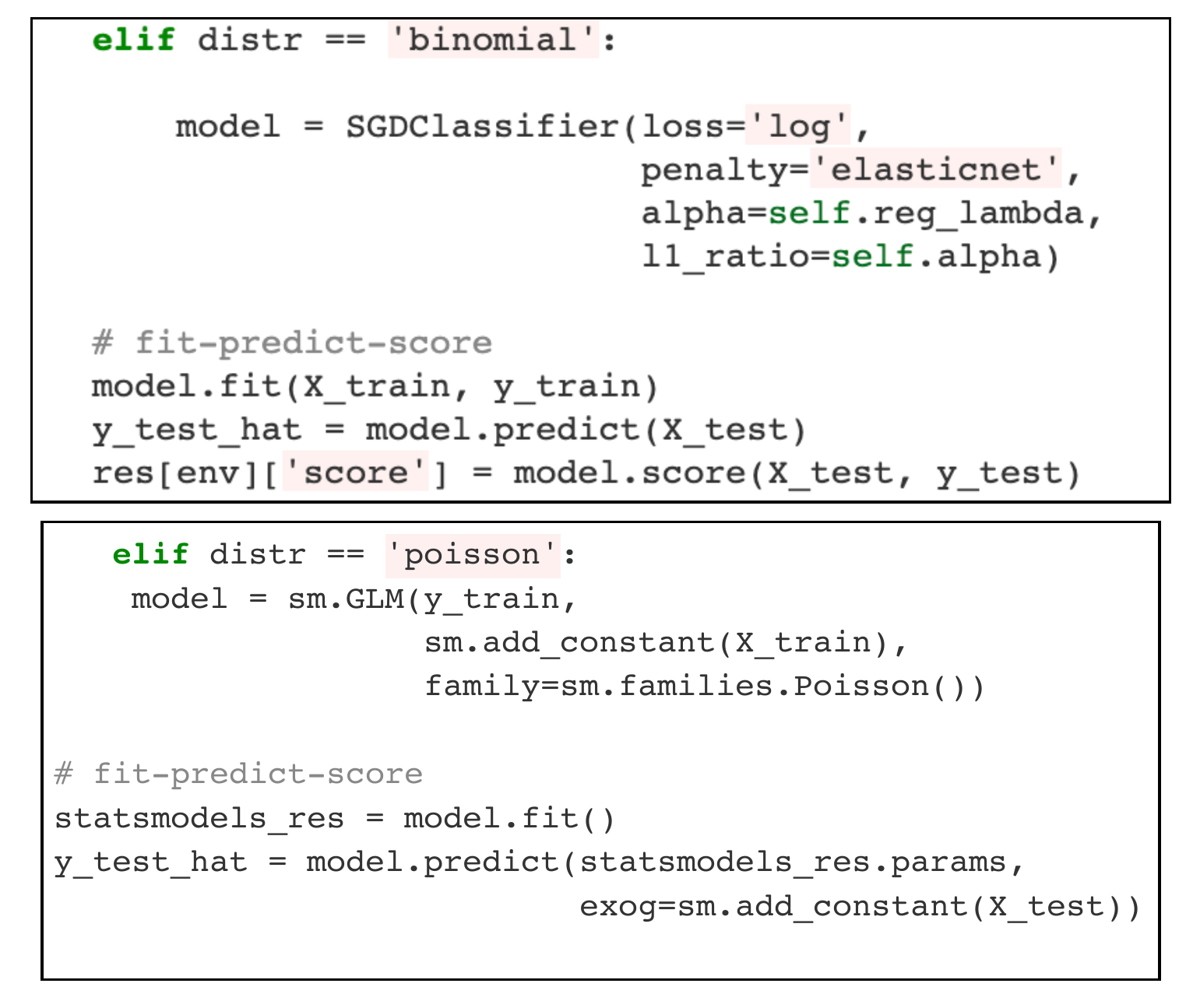}
  \end{center}
  \caption{Usage in code}
  \label{code_usage}
\end{figure}

Code understanding is an increasingly important application of AI, with over 100 papers targeting the area in the last year alone\footnote{\url{https://ml4code.github.io/papers.html}}. 
 Much research in this area has focused on understanding code from abstract representations of the program such as Abstract Syntax Trees (ASTs) and program flow.  However, there has been little emphasis in utilizing important semantics about code buried in textual artifacts, such as documentation or forum discussions.  Extracting such information can significantly enrich code representations.  As an example, Figure~\ref{code_usage} shows a program where the classes \textit{GLM} and \textit{SGDClassifier} are being used.  If one could enrich the representation of the two classes in the code with their key features from text, we would understand that both represent linear models, and hence both code snippets perform similar functions.  
 
 To enrich code with textual information, we need to be able to summarize textual information about classes and functions into vector representations.  Pre-trained language models are an obvious choice, but we currently do not know how applicable they are to text about code, given the specialized language of the programming domain.  We need a set of code-related downstream tasks to evaluate these models, just as GLUE \cite{wang-etal-2018-glue} and SuperGLUE \cite{SuperGLUE} have been used extensively to further language model development in the natural language understanding domain. CodeXGLUE \cite{CodeXGLUE} provides a suite of tasks but only a single task in it is related to textual code artifacts; it is translation of documentation about code from one natural language to another.  To our knowledge, we know of no other tasks that focus on relations between textual artifacts about code.  This paper attempts to fill this gap.
 
 We have three goals in this paper: (a) design a suite of tasks we refer to as BLANCA (Benchmarks for LANguage models on Coding Artifacts) that can be used to train language models about the \textit{semantics of code}, (b) evaluate whether existing models, fine-tuned for different aspects of natural language processing or different code oriented corpora, can perform well on these tasks, and (c) establish whether these tasks can be used to build better models for code understanding.

To construct these tasks, we relied on existing annotations in large public repositories such as GitHub (for code), StackOverflow, StackExchange and code documentation (for text about code).  We exploited an integration of these sources in an open source dataset  \cite{abdelaziz2020graph4code} to define the following five tasks focused on text about code:
\begin{itemize}
   \item \textbf{Forum Answer Ranking (\textit{R})}.  Some answers on forums have many votes or are selected as the best relative to others. Can language models predict the best answers?
   \item \textbf{Forum Link Prediction (\textit{L})}.  Users of forum posts often point to other similar posts, which reflect semantically related posts compared to random pairs.  Can language models predict links?
    \item \textbf{Forum to Class Prediction (\textit{F})}.  Key features of classes or functions often get discussed in forum posts.  Do language models discriminate related posts and class documentation from unrelated ones?
    \item \textbf{Class Hierarchy Distance Prediction (\textit{H})}. Code is often organized into class hierarchies.  Do embedding distances from language models reflect class distances in the hierarchy?
    \item \textbf{Class Usage Prediction (\textit{U})}.  Similar code is often used in similar ways.  Are embedding distances smaller for documentation about classes that are used similarly, and larger for dissimilar ones?
\end{itemize}


We compare performance on these tasks for seven language models, chosen for differences in architecture, training tasks, and corpora, as outlined in Section~\ref{sec:models}. Our main findings are as follows:
 \begin{itemize}
     \item Out of the box, language models trained on general corpora perform reasonably well on most BLANCA tasks, compared to models trained on code specific corpora such as CodeBERT \cite{feng2020codebert} or BERTOverflow \cite{tabassum2020code}, attesting to the generality of these models. 
     \item However, on every task, fine tuning on code specific models resulted in significant boost in performance, highlighting the usefulness of BLANCA tasks for building better language models.
     \item Multi-task training produced better performance on many BLANCA tasks, suggesting the tasks do help models learn code semantics that transfers across tasks.  
 \end{itemize}
 To aid further research in code understanding, the code, datasets and the fine-tuned models are publicly available\footnote{\url{https://github.com/wala/blanca}} under an open source license (Eclipse for the code and Creative Common with Attribution for the data), and we hope they prove useful to the code understanding community to enrich representations of programs with textual information about classes and functions.  
 
\section{Related Work}
There have been numerous benchmarks for code summarization or generation of code from natural language, and hence they have focused on collecting code and textual documentation that characterize the code.  For these tasks, most have used the approach of generating code and its associated documentation strings, e.g., \cite{leclair-mcmillan-2019-recommendations}, \cite{movshovitz-attias-cohen-2013-natural}.  Similarly, code and corresponding textual documentation have been used for numerous tasks involving searching for code, e.g., \cite{li2019neural}, \cite{husain2019codesearchnet}, or searching for posts given code, e.g., \cite{10.1145/2597073.2597077}.  

While such benchmarks are useful for joint embeddings of code and their associated text, they are restricted to tasks around code summarization, code generation, comment generation or code search; i.e., they do not directly help with the evaluation of language models for textual artifacts about code. Furthermore, most of the datasets in the literature do not correlate textual artifacts around code with code usage, with the exception of \cite{ijcai2018-314}, which does link the generation of API sequence information from their usage in code to the problem of code summarization. The work in \cite{7962378} connects code on GitHub to StackOverflow posts, but the latter dataset is not available.  Again, their datasets are targeted to the task of code summarization, and code search respectively.  Similarly StackOverflow posts have been used for tasks such as answer summarization \cite{10.1145/3338906.3341186}, program repair \cite{8330202} or generating code, e.g., \cite{8330202}. Finding directly related or duplicate posts is a recent task and dataset proposed in \cite{Shirani2019QuestionRO}, but there is no evaluation of any language model in that work.

Recently, \cite{tabassum2020code} provided a BERTOverflow model for an in-domain representation of text about code.  BERTOverflow is trained on 152 million StackOverflow questions over a BERT architecture, and has been fined tuned for software named entity recognition (e.g., finding mentions of operating systems in text). We use BERTOverflow as one of the models for the BLANCA tasks. There has also been work building structural language models from the abstract syntax trees, e.g., \cite{ANY-Code}, which is clearly a related task, but the focus is once again on code.  CodeXGLUE \cite{CodeXGLUE} provides a novel text-text benchmark which involves translation of documentation about code from one language to another, but that is arguably closer to natural language processing than code.  

Thus, to the best of our knowledge, no work so far has examined how language models perform on a set of code related tasks for textual program artifacts, nor has there been much emphasis on building benchmarks to build better text representations for code understanding. BLANCA is built to address this gap.

\section{Models}
\label{sec:models}
In choosing models for our experimentation, we needed language models to encode paragraphs in either class documentation or posts.  We relied largely on the sentence transformers library \cite{reimers-2019-sentence-bert}, which provides a wide range of transformer models that have been fine-tuned for tasks such as information retrieval, paraphrase detection, and sentence similarity detection. These models have been shown to be effective in sentence and paragraph encoding style tasks.  We also chose models with a different base, such as BERT \cite{devlin-etal-2019-bert}, XLM-RoBERTa \cite{DBLP:journals/corr/abs-1911-02116} and DistilBERT\cite{sanh2020distilbert}.  We also added a non-transformer style model (Google's Universal Sentence Encoder \cite{cer-etal-2018-universal}), and models fine-tuned on StackOverflow posts (BERTOverflow \cite{tabassum2020code}) and code documentation (CodeBERT  \cite{feng2020codebert}) to see if domain-specific training is helpful.  We did not consider models, such as CuBERT \cite{cubert}, designed only for code, and not text about code. The reason is that  cuBERT's vocabulary is based on  programming language tokens for Java or Python, which is only partially useful for text about code.  Table~\ref{table:models} shows the types of base models used in our evaluation, using the names from the sentence-transformers (SBERT\footnote{\url{https://github.com/UKPLab/sentence-transformers}}) library.  

\begin{table}[]
{\small
\centering
\begin{tabular}{ll}
\toprule
 Model name & Fine-tuning task \\
\midrule         
Universal Sentence Encoder$^\ddagger$ & N/A \\
BERT-NLI$^\dagger$ & Sentence similarity    \\
DistilBERT-paraphrasing$^\dagger$ & Paraphrase detection  \\
xlm-r-paraphrase-v1$^\dagger$ & Paraphrase detection \\ 
mmsmarco-DistilRoBERTa$^\dagger$ & Information Retrieval \\
BERTOverflow$^\dagger$ & StackOverflow/NER \\
CodeBERT-mlm$^\dagger$ & NL-PL pairs in 6 languages \\
\bottomrule
\end{tabular}
\caption{Models used as baselines. Sources were tensorflow-hub, and SBERT. bert-base-nli-stsb-mean-tokens, distilroberta-base-paraphrase-v1, xlm-r-distilroberta-base-paraphrase-v1, msmarco-distilroberta-base-v2 and microsoft/codebert-base-mlm are their corresponding names in SBERT. }
\label{table:models}}
\end{table}

 We also tested if fine-tuning on each task would enhance performance, to establish whether the tasks can be used to build a better language model.  For fine-tuning, we started either with BERTOverflow or CodeBERT, with the assumption that an in-domain representation would provide some advantage.  We also examined whether multi-task training would improve performance, to see if better models could be built from using a combination of BLANCA tasks.

\section{Tasks}
All our datasets describe code artifacts in Python, and are derived from Graph4Code, which links 1.3 million programs of Python code to associated posts and class-documentation \cite{abdelaziz2020graph4code}.
For multi-task  fine tuning, we report, for each task, the model with the best performance, and we outline its characteristics.  In Section~\ref{sec:multitask}, we discuss more general findings for multi-task training. Performance on tasks is encoded as follows in tables: (1) Forum Answer Ranking (R), (2) Forum Link Prediction (L), (3) Forum Class Prediction (F), (4) Class Hierarchy Prediction (H), and (5) Class Usage Prediction (U). Table \ref{data_stats} lists each BLANCA task and the corresponding train/test data sizes.

 \begin{table*}[]
\centering
\begin{tabular}{llll}
\toprule
                                    & Data type               & Train      & Test      \\
\midrule
Forum Answer Ranking                & Question-answer pairs   & 450,000    & 50,000    \\
Forum Link Prediction               & Question-Question pairs & 23,516     & 5,854     \\
Forum to Class Prediction           & Question-Class pairs        & 11,488     & 1,275     \\
Class Hierarchy  Prediction & Class-Class pairs       & 16,215,400 & 1,801,716 \\
Class Usage Prediction              & Class-Class pairs       & 75,862     & 8,439 \\
\bottomrule
\end{tabular}
\caption{BLANCA's tasks and datasets statistics}
\label{data_stats}
\end{table*}


\subsubsection{Dataset Annotation Quality} 
Two of BLANCA tasks are based on manually curated datasets by millions of users such as ranking answers in StackOverflow forums (Forum Answer Ranking) and manually linking similar posts (Forum Link Prediction).
These data are high quality, in the sense that they are crowd annotated by humans, which is how most gold standards get constructed.  Class Hierarchy and Class Usage Prediction tasks are both based on objective properties of code artifacts (class hierarchy and similarities among classes in terms of their methods, respectively), so once again, the issues of data quality do not arise.  The only task where we did not have explicit human labeling for every example is Forum to Class Prediction. In this task, we relied on heuristics to automatically label the data. Furthermore, to assure quality, we performed a manual evaluation of a sample with three human annotators (see Section \ref{forum2class_sec} for details).

\subsubsection{Hyperparameter Search for Finetuning} 

We started with the default parameters of our base models; CodeBERT and BERTOverflow. We also tried to use Population Based Training from
RayTune\footnote{https://docs.ray.io/en/latest/tune/index.html} to perform hyper-parameter search for the Forum Answer Ranking (R) and Forum Link Prediction (L) tasks. However, we did not get better performance compared to using the default parameters from the corresponding base models.  

We describe below how we formulated each task, the dataset definition process and the performance of various language models on it.  

\subsection{Forum Answer Ranking (R)}

\subsubsection{Task Description}
StackOverflow and StackExchange contain questions and answers. Accepted answers are manually annotated and most answers have a vote count.  The core task here is to predict the best answer to each question, and order the answers by their popularity. 

\subsubsection{Dataset}
We generated a dataset of 500K questions such that each question comes with at least three answers. The average number of answers per question in this dataset is 4.9 answers, and the average number of votes per question is 23.5 and per answer is 12.74. The train and test tasks were split 90-10, so the train set had 450,000 questions and test had 50,000 questions.  To build the fine tuning model, we modeled this as a task similar to training on the Semantic Textual Similarity Benchmark (STSB) adopted by SBERT.  Each answer was ranked according to popularity, and ties were broken by adding only one of the answers that were tied.  The ranks were then converted to a score between 0 (worst rank) and 1 (best rank), with a cosine similarity loss, and an embedding similarity evaluator from the SBERT library.  Fine tuning was performed on BERTOverflow and CodeBERT models, with the 90\% of training data for training, 10\% of the training data for validation, for 10 epochs.  

\subsubsection{Evaluation}
To capture how well the embeddings of different language models identified the ranking of answers, we computed the cosine distances between the question embedding and the embedding of each of the answers, and ranked answers by nearest in cosine distance to furthest.  We report standard information retrieval metrics of average Mean Reciprocal Rank (MRR) and average Normalized Discounted Cumulative Gain (NDCG) on this predicted ranking.   

Table~\ref{tab:ranking} shows that most language models do reasonably well on this task, which is not surprising because text in forum posts is mostly natural language.  Surprisingly though, there is no benefit for the base BERTOverflow model that has been tuned on StackOverflow posts compared to the rest of non-finetuned models. However, fine-tuned BERTOverflow does much better, which is consistent with our hypothesis that it is possible to use these tasks for building better language models.  Across many tasks, fine-tuning on BERTOverflow produced better performance than fine-tuning on CodeBERT, which suggests that  forum discussions contain in most cases, the right mixture of explanations in natural language along with code.  Moreover, the best performance was achieved with multi-task finetuning (RFLHU-BERTOverflow), which suggests that use of multiple BLANCA tasks builds better language models for textual code artifacts.

\begin{table}[]
{\small
\centering
\begin{tabular}{lll}
\toprule
        & MRR                      & NDCG                                   \\
\midrule    


DistilBERT-paraphrasing & 0.5937	 \tiny{(.001)}	& 0.8393	 \tiny{(.001)} 	\\
BERT-NLI &	0.5972 \tiny{(.001)}	&	0.8407 \tiny{(.001)} 	\\
msmarco-DistilRoBERTa &	 0.5992 \tiny{(.001)}	&0.8427	 \tiny{(.001)} 	\\
xlm-r-paraphrase-v1 & 0.5977 \tiny{(.001)} 	& 0.8411	 \tiny{(.001)}  \\
USE &0.6114	 \tiny{(.001)}	&0.8483	 \tiny{(.001)} \\	
BERTOverflow &	0.5910 \tiny{(.001)}	&	0.8375 \tiny{(.001)}\\	
CodeBERT & 0.5926\tiny{(.001)} & 0.8375 \tiny{(.001)} \\
\hline
FT-BERTOverflow &	0.6743 \tiny{(.001)} & 0.8823\tiny{(.001)} \\
FT-CodeBERT &  0.6671\tiny{(.001)} &  0.8790\tiny{(.001)} \\
\hline
RFLHU-BERTOverflow &	\bf 0.6879 \tiny{(.001)} & \bf 0.8893	 \tiny{(.001)} \\

\bottomrule

\end{tabular}
\caption{Performance of language models on forum answer ranking (R). The numbers in parentheses are the standard errors of the sample mean. FT represents fine tuning on R alone, RFLHU-BERTOverflow is the best multi-task training model. }
\label{tab:ranking}}

\end{table}



\subsection{Forum Link Prediction (L)}
\subsubsection{Task Description}
Forum posts with links to one another are usually related compared to unlinked posts; we investigate if language models place such related post pairs closer in vector space.  We focus on embedding distance because it is a more direct metric for assessing the quality of the embedding rather than classification accuracy.

\subsubsection{Dataset}
For this task, we generated 23,516 pairs of posts for training (11,758 positive and 11,758 negative), 5,854 pairs (2,727 positive and 2,727 negative) for testing.  Fine-tuning was set up as a classification task in SBERT, with the use of contrastive loss along with a binary classification evaluator from the SBERT library.  All other training details were similar to the forum answer ranking task. 

Relevant to this task, \cite{Shirani2019QuestionRO} recently introduced a similar benchmark for predicting relatedness in StackOverflow posts focused on Java code, as opposed to our dataset which is language agnostic. Their dataset contains 300K of linked pairs categorized into 1) duplicates: questions in StackOverflow marked by moderators as duplicates, 2) direct: explicitly linked posts, 3) indirectly or transitively connected posts through a direct or a duplicate link and 4) isolated or unlinked posts. Direct and isolated links are similar to our positive and negative examples. We evaluate all our models' ability to differentiate these link types. Note that we did not use this data for fine-tuning a model which discriminates the different categories; but one might expect direct links and duplicates to be closer in embedding distance, and isolated links to be the furthest, with indirect links in the middle.  \citet{Shirani2019QuestionRO} did not evaluate this with any of the language models, so we examine whether these categories of relatedness of posts is reflected in embeddings of pre-trained models.


\begin{table}[]
\centering
{\small
\begin{tabular}{lrrr}
\toprule
Model        & Linked      & Unlinked    & T                             \\
\midrule    

DistilBERT-paraphrasing &	0.38	&	0.71 & 112.49	\\
BERT-NLI &	0.31	&	0.53 & 74.92	\\
msmarco-DistilRoBERTa &	0.34	&	0.74 & 110.42	\\
xlm-r-paraphrase-v1 &	0.37	&	0.70 & 105.02 \\
USE &	0.34	&	0.74  & 142.04 \\	
BERTOverflow &	0.20	&	0.31 & 59.52 \\
CodeBERT & 0.03 & 0.04 & 19.39 \\
\hline
FT-BERTOverflow &	0.09  &	0.52  & 180.42 \\
FT-CodeBERT & 0.08 & 0.50 & 147.21 \\
\hline
RFLHU-BERTOverflow & \textbf{0.08} & \textbf{0.58} & \textbf{198.10} \\

\bottomrule
\end{tabular}}

\caption{Cosine distance between linked and unlinked posts (L). FT represents fine-tuning on L alone.}
\label{tab:posts}

\end{table}

\subsubsection{Evaluation}
\begin{figure}
    \includegraphics[width=.9\linewidth]{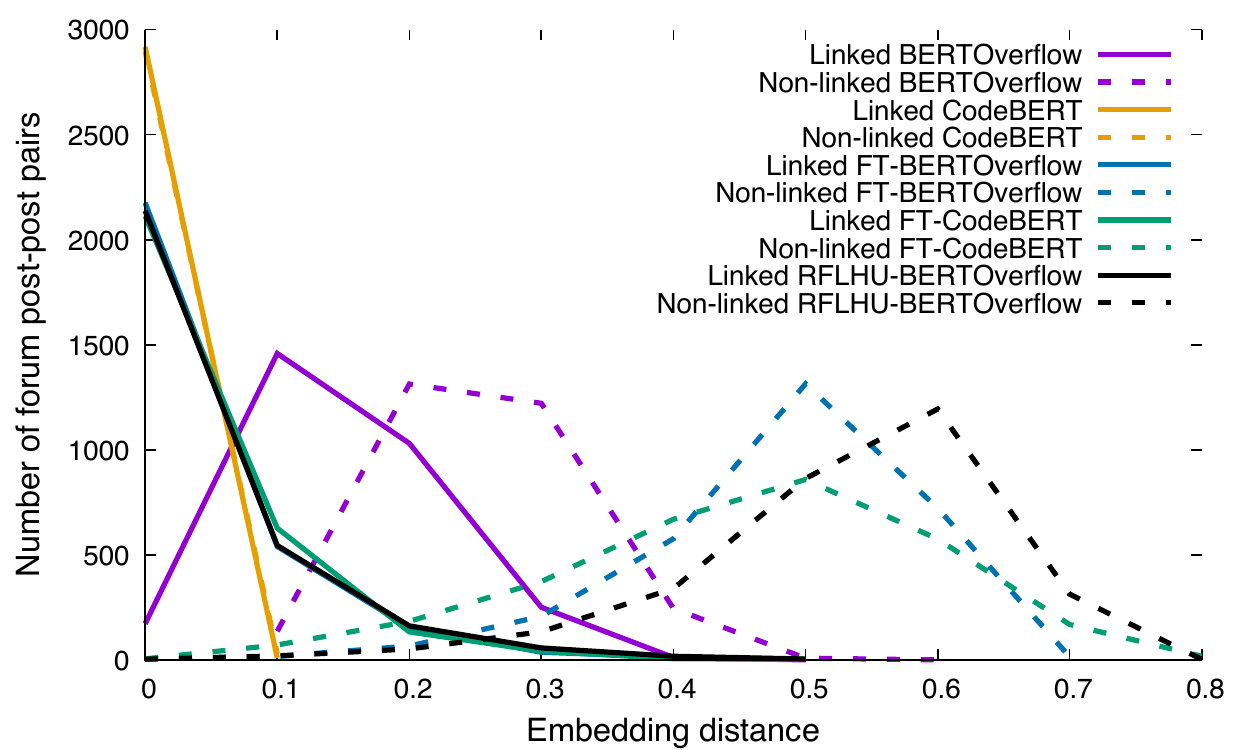}
  \caption{Linked versus unlinked pair distances for all models (L).}
  \label{linked_post_dists}
\end{figure}

As shown in Table~\ref{tab:posts}, all language models showed a statistically significant difference ($p \leq .01$) on independent sample t-tests between linked and unlinked posts. 
BERTOverflow with fine-tuning (both versions tuned on L only and RFLHU) performing the best in terms of pulling apart linked and unlinked posts.  We note that the size of T value normalizes the distance between linked and unlinked posts by their variance; that is, the T value captures not only the average distance but also the separation between the two distributions.  Our focus then is on the absolute value of that separation as provided by the T value.  Figure~\ref{linked_post_dists} shows this visually. 
We note that BERTOverflow and CodeBERT  as base models discriminated least between linked and unlinked posts, but fine-tuning clearly helped greatly.  
This is evident in the solid and dashed lines for RFLHU-BERTOverflow  where it shows little overlap between linked and unlinked posts. 

\begin{figure}
    \centering
    \includegraphics[width=.9\linewidth]{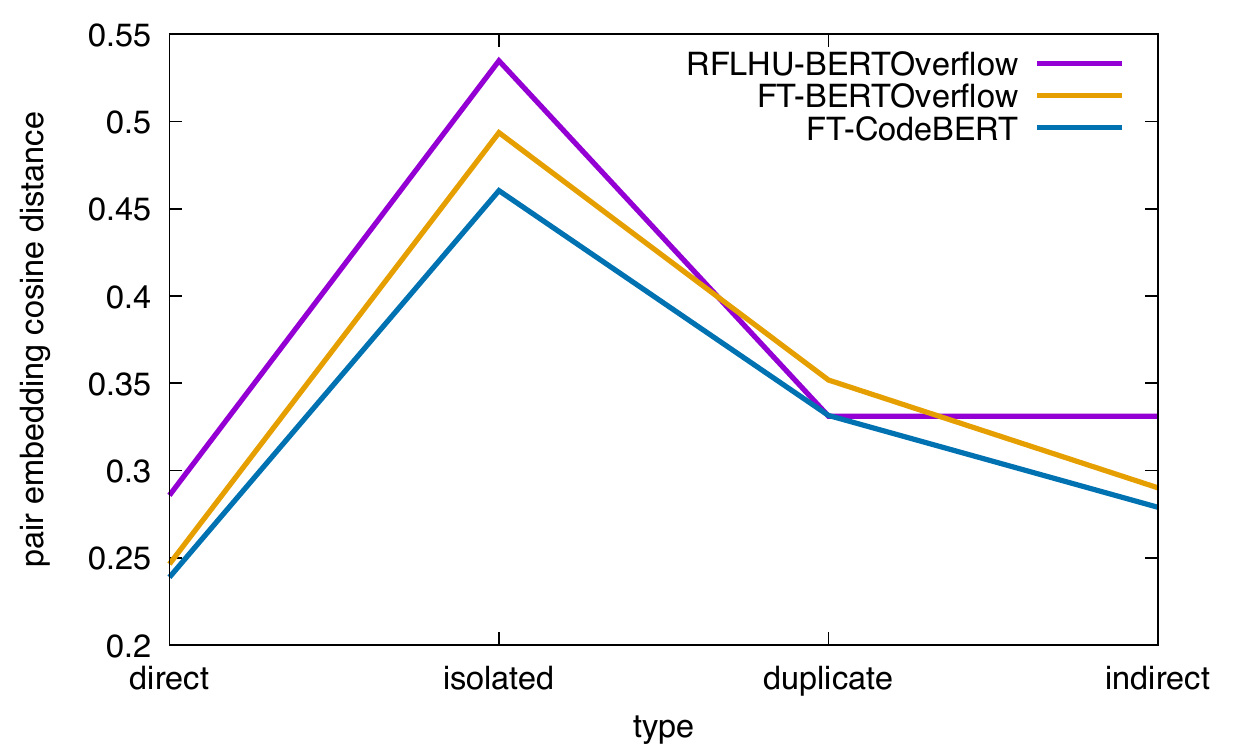}

  \caption{Direct, indirect, duplicate and isolated pair distances for all models.}
  \label{related_questions}
\end{figure}

Figure~\ref{related_questions} shows the results of  a variety of language models for question relatedness variant of this task~\cite{Shirani2019QuestionRO}. We ensured that none of ~\citet{Shirani2019QuestionRO}'s test set examples were used in our training set.  Across all models, directly related questions are closest in embedding space followed by indirectly related questions. Questions marked duplicate posts were similar to the indirect questions only in the RFLHU model, which seemed to be picking up relatedness in both indirect and duplicate questions.  We note that questions marked duplicates in forums are only duplicates at a level of coding abstraction.  For example, the two questions ``How to return multiple objects from a Java method?" and  ``Java how to return two variables?" are a duplicate pair. Although the two questions talk about the same problem, the discussions and even the solutions are different. Therefore, cosine similarity between them is not as close as one would expect. Finally, isolated question pairs are the most distant compared to all other pairs across all models (all differences from isolated pairs to direct, indirect and duplicate pairs were statistically significant at the .01 level).  Multi-task fine-tuning (RFLHU-BERTOverflow) clearly helped the best in getting semantically related posts closer and pulling apart the unrelated ones. 

\subsection{Forum to Class Prediction (F)}
\label{forum2class_sec}
\subsubsection{Task Description}
Forum posts often describe specific code artifacts in text, where they discuss key features of a class or a function.  A key question is whether a model can predict if a post about a class and documentation of the same class are related. 


\subsubsection{Dataset}
In order to find posts that were more focused on discussions of a specific class or function's features, we queried an ElasticSearch index of posts with a query per class as in Graph4Code \cite{abdelaziz2020graph4code}, insisting that the class and its package be both mentioned in the question. These constituted our positive class-post examples.  
For negative examples, we chose hard negatives, requiring that both  class name and its package not be mentioned anywhere within the question and its answers; but nevertheless the post matched either class or package names. 
To ensure the quality of this data, we asked 3 annotators to label a random sample of 100 examples; 50 positive and 50 negative. 
This manual inspection revealed that negatives were in fact negatives, in the sense that even if the class was mentioned, it was usually, from a different package, or very often from different programming languages (e.g. Java, Javascript, etc). 
The average hit and miss rates from the three annotators were 96.7\% and 3.3\%, respectively. 
In this task, we created 8,827 negative examples and 2,661 positives for training, and 980 negative examples and 295 positive examples for testing. Fine-tuning the model was analogous to the forum link prediction task.




\begin{table}[]
\centering
{\small
\begin{tabular}{lrrr}
\toprule
Model        & Related      & Unrelated    & T                             \\
\midrule    

DistilBERT-paraphrasing &0.55		&0.68	 & 16.61	\\
BERT-NLI &	0.45	&0.60	 & 14.73	\\
msmarco-DistilRoBERTa &	0.45	&0.66	 & 20.37	\\
xlm-r-paraphrase-v1 &	0.53	&0.67	 & 17.15 \\
USE &	0.53	&0.74	  &  20.67\\	
BERTOverflow &	0.33	&0.47	 & 18.31 \\
CodeBERT &  0.06& 0.09 & 12.23 \\
\hline
FT-BERTOverflow & 0.07  &0.77	  & 46.88  \\
FT-CodeBERT & 0.08 &0.82  & 50.07 \\
\hline
RFLHU-CodeBERT & \bf 0.11 & \bf 0.66 &  \bf 53.98\\

\bottomrule
\end{tabular}}
\caption{Cosine distance between documentation-post pairs (F) that are related and unrelated. FT represents fine-tuning on F alone.}
\label{tab:class_posts}
\end{table}

\subsubsection{Evaluation}
\begin{figure}
\centering
\includegraphics[width=.9\linewidth]{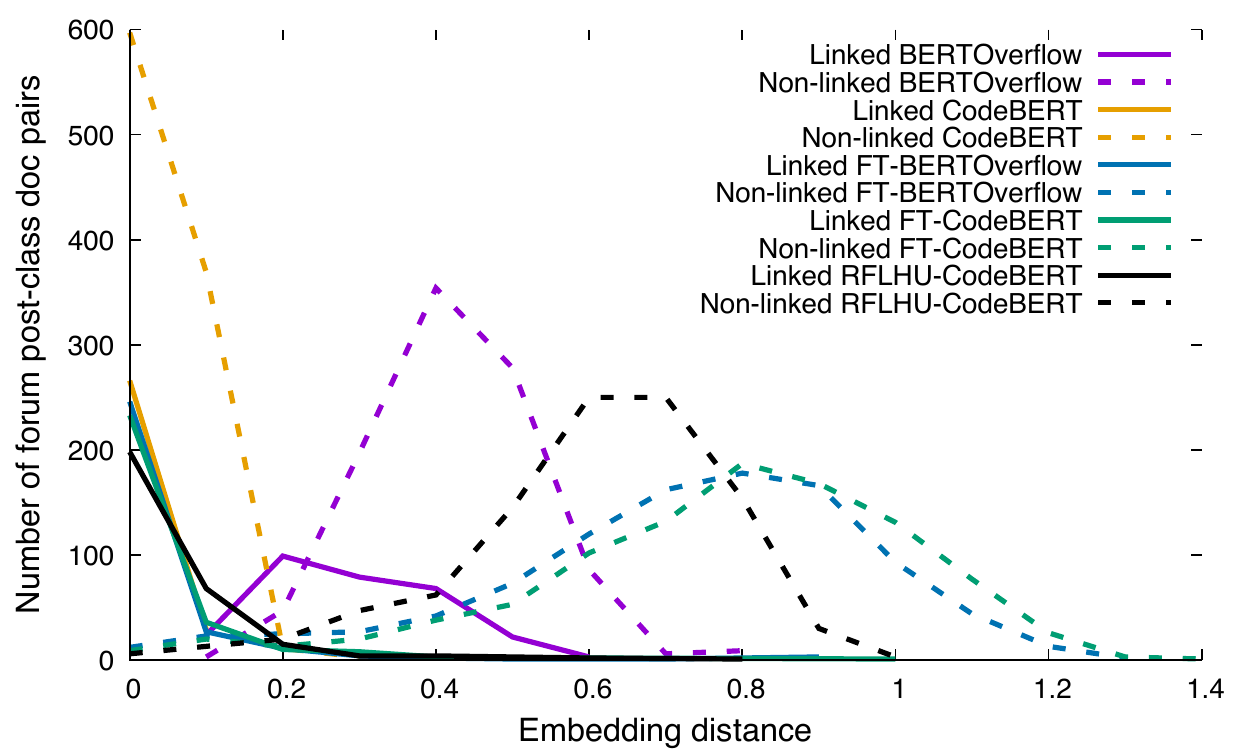}

  \caption{Related and unrelated documentation-post  pair distances for some fine-tuned and non-fine-tuned models (F).}
  \label{class_forum_dists}
\end{figure}

As shown in Table~\ref{tab:class_posts}, all language models showed a statistically significant difference ($p \leq .01$) on independent sample t-tests between positive and negative class-post examples. Again, fine-tuning helps improving the performance on this task significantly; e.g. single task tuning of FT-CodeBERT vs. CodeBERT and FT-BERTOverflow compared to BERTOverflow. Fine-tuning on multiple tasks, e.g. RFLHU-CodeBERT, gave better performance compared to the single-task tuned models, FT-CodeBERT and FT-BERTOverflow. As shown in Figure~\ref{class_forum_dists}, the distance was greatly enhanced by fine-tuning e.g. BERTOverflow vs. fine-tuned RFLHU-BERTOverflow.

\subsection{Class Hierarchy Distance Prediction (H)}
\subsubsection{Task Description}
Semantically related classes tend to be linked by developers in a class hierarchy, so its reasonable to ask if neural embeddings of related classes cluster closer together in a class hierarchy.  We structured this as a class distance prediction task, with class distances ranging from 1 to 10.

\subsubsection{Dataset}
We collected the documentation associated with 257,655 classes in Graph4Code \cite{abdelaziz2020graph4code}, but many of these represent different names that resolve to the same class.  We aliased the classes to its canonical version by loading the class dynamically to obtain its runtime name, and added in classes that we could not load for some reason, which resulted in 90,464 classes.

\begin{figure}
  \begin{center}
  \includegraphics[width=.8\linewidth]{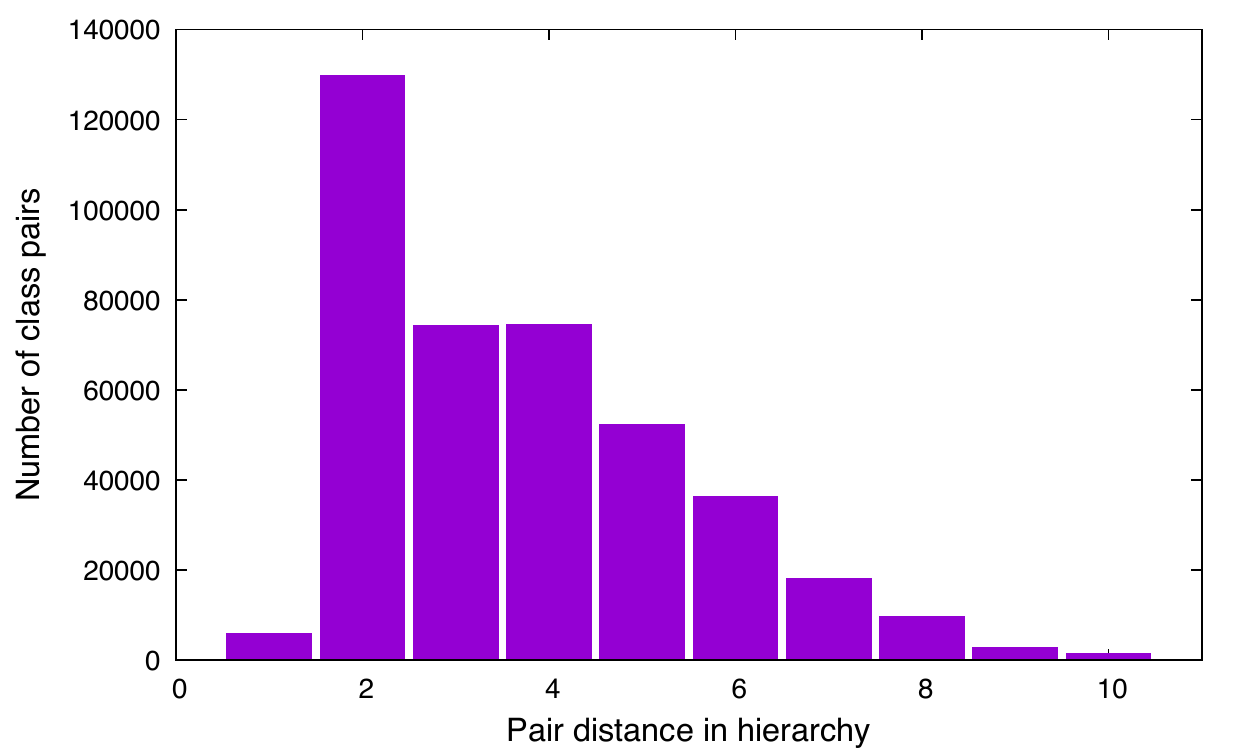}
  \end{center}
  \caption{Number of class-pairs by class distance.}
  \label{hierarchy_distance_distribution}
\end{figure}

To get classes related by distance, we created an undirected graph of class to superclass relations for every module, being careful not to add edges from any class to the class \texttt{object}. For each module graph, we computed distances between every pair of classes using an all pairs shortest paths algorithm.  We eliminated pairs with distances greater than 10, and this resulted in a set of pairs that we split randomly such that 16,215,400 million pairs of classes were in train, and 1,801,716 million pairs were in test.  For fine-tuning, we structured this similar to the forum ranking task, with distances translated to scores between 0 (least related) and 1 (most related), and we used cosine similarity loss, coupled with a embedding similarity evaluator from SBERT.  Training on 16.2 million pairs was computationally expensive so we trained it on a random sample of 100,000 training examples, 10,000 of which was used for validation. Figure~\ref{hierarchy_distance_distribution} shows the distribution of embedding distances for each class distance (1-10) to show the dataset characteristics.  

\begin{figure}
\centering
\includegraphics[width=.85\linewidth]{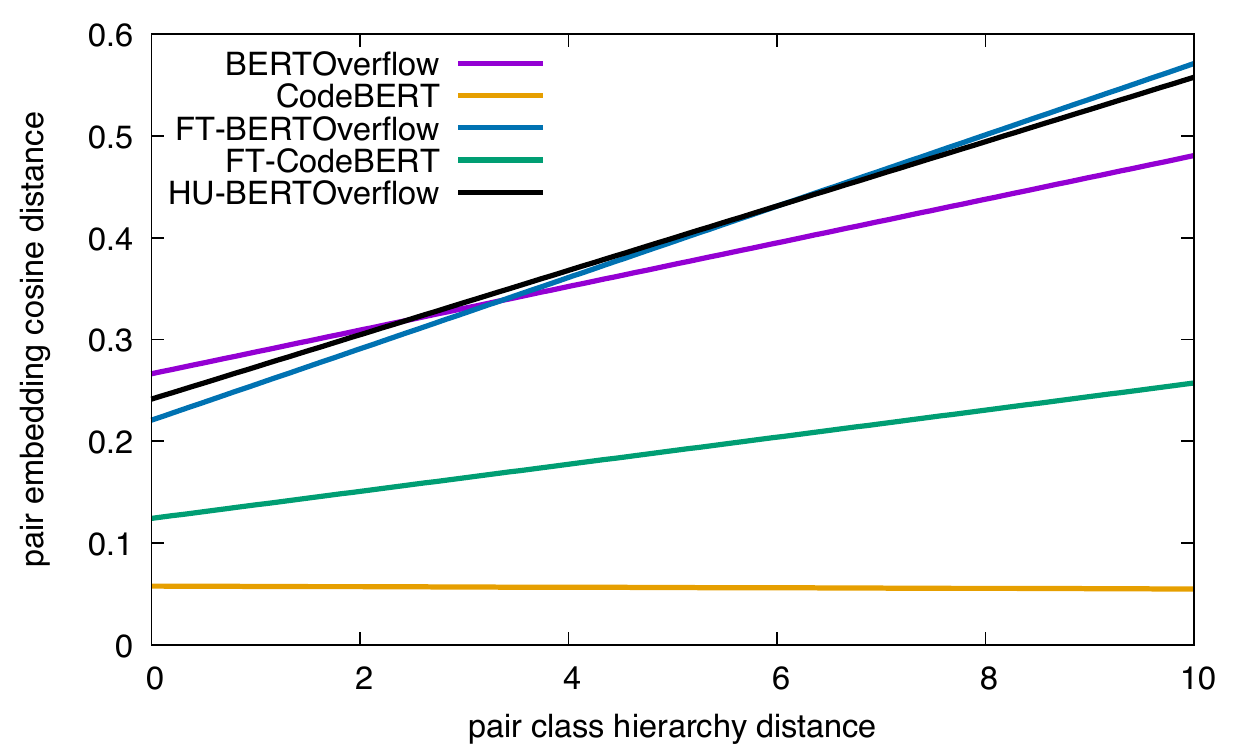}

  \caption{Prediction of embedding distance from class distance (H) for all models. Standard error of regression was less than 0.0002 for all models.}
  \label{docstring_distance}
\end{figure}

\subsubsection{Evaluation}
Since this is a regression task, we evaluated the Pearson $r$ correlation, which as shown in Table~\ref{tab:correlation_hierarchy} varied from 0.17 (for BERTOverflow) to 0.34 (for Fine-tuned-BERTOverflow); all are statistically significant at  $p \leq 0.01$.  Regression for each model is shown in Figure~\ref{docstring_distance}.  The improvement from fine-tuning for BERTOverflow showed that the task is useful for building better embeddings.  Some other models showed reasonable performance with no tuning (xlm-r-paraphrase at 0.27, and USE at 0.28 respectively), so there is clearly a room to improve these different base models as well, but we leave that issue for future work, since our goal is more on task development rather than building better models.

\begin{table}[]
\centering
{\small
\begin{tabular}{lr}
\toprule
Model       & Pearson $r$                                                       \\
\midrule    

DistilBERT-paraphrasing &	0.26 \\
BERT-NLI &	0.20 \\
msmarco-DistilRoBERTa &	0.23	\\
xlm-r-paraphrase-v1 & 0.27 \\
USE &	0.28 \\	
BERTOverflow &	0.17 \\	
CodeBERT & -0.01 \\
\hline
FT-BERTOverflow & \textbf{0.34} \\
FT-CodeBERT & 0.24 \\
\hline
HU-BERTOverflow & 0.29 \\
\bottomrule
\end{tabular}}
\caption{Correlation of class hierarchy (H) distance to embedding distance by model. FT represents fine-tuning on H alone.}
\label{tab:correlation_hierarchy}
\end{table}

\subsection{Class Usage Prediction (U)}
\subsubsection{Task Description}
GitHub contains millions of programs, where classes are used in code to achieve some purpose.  Classes that are used in the same way; i.e., same set of methods get invoked on them, might be expected to be rated as more similar than classes that do not share any methods.  We structured this as a similarity rating task.   

\subsubsection{Dataset}
To construct this dataset, we used the Graph4Code knowledge graph \cite{abdelaziz2020graph4code} which has data flow graphs for 1.3 million GitHub programs.  Dataflow tracks the flow of data through return values and parameters within a program.  As an example, for the program snippet shown in Figure~\ref{code_usage}, dataflow would show that \texttt{fit} and \texttt{predict} calls occur on objects returned by calls to the constructors of \texttt{SGDClassifier} and \texttt{GLM}.  In this example, \texttt{SGDClassifier} shares 2 methods (denoted as $M$) with 1 class (denoted as $C$), which in this instance is \texttt{GLM}).  The classes are similar, in the sense that they both share the same methods in usage, but the degree of similarity is dependent on the number of shared methods ($M$), and the number of classes that have the same methods ($C$).  The smaller the $C$, the more likely it is that a pair is similar, and the larger the $M$ the more likely it is that the pair is similar.  To capture both dimensions of similarity into a single distance metric for learning, we defined an `ideal' class pair in terms of our data - that is a vector with $[max(M), min(C)]$.  We used the Euclidean distance of each class pair from this ideal vector as the dissimilarity metric.  Given a pair, the task then is to predict if the classes were similar or distant based on their usage.

The train task contains 75,862 class pairs, and the test task contains 8,439 pairs, with an average distance of 312.21 for train pairs and an average distance of 312.12 for test pairs, suggesting the two had similar characteristics.  The fine-tuning task was modeled the same as the class hierarchy prediction task.


\subsubsection{Evaluation}
\begin{figure}
\centering
    \includegraphics[width=.85\linewidth]{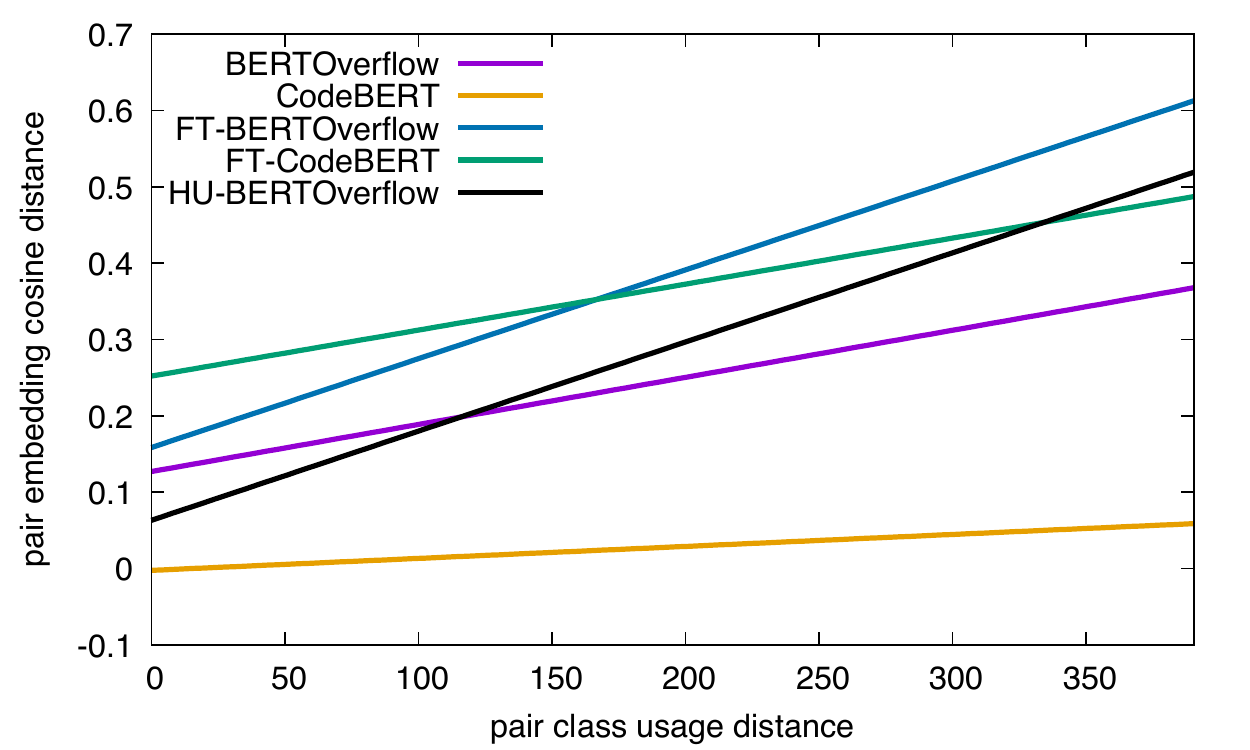}
  \caption{Prediction of embedding distance by class usage (U) similarity for all models. Standard error of regression was less than 1.0e-4 for all models.}
  \label{usage_results}
\end{figure}

 We frame this task as a regression task and evaluate the Pearson $r$ correlation once again for all models, which varied from 0.17 (for BERT-NLI) to 0.61 (for HU-BERTOverflow) as shown in Table~\ref{tab:usage}; all results are statistically significant at  $p \leq 0.01$.  The improvement from fine-tuning for BERTOverflow (0.37 to 0.52) also shows the task is useful for building better embeddings.  Using hierarchy task as well with HU-BERTOverflow further improved performance to 0.61. This was not the case though for CodeBERT models with and without fine-tuning where its performance dropped to 0.30 from 0.33. We also show in Figure~\ref{usage_results} the effectiveness of usage distance as a predictor of cosine embedding distance.  

\begin{table}[]
\centering
{\small
\begin{tabular}{lr}
\toprule
Model       & Pearson $r$                                                       \\
\midrule    
DistilBERT-paraphrasing &	0.35  \\
BERT-NLI &	0.17 \\
msmarco-DistilRoBERTa &	0.34 	\\
xlm-r-paraphrase-v1 & 0.36  \\
USE &	0.41 \\	
BERTOverflow &	0.37 \\	
CodeBERT & 0.33 \\
\hline
FT-BERTOverflow & 0.52 \\
FT-CodeBERT & 0.30 \\
\hline
HU-BERTOverflow & \textbf{0.61} \\
\bottomrule
\end{tabular}}

\caption{Correlation of class usage similarity (U) with embedding distance by model. FT represents fine-tuning on U alone.}
\label{tab:usage}
\end{table}

\subsection{Multi-Task Training}
\label{sec:multitask}
We focus our multi-task training discussion on BERTOverflow, because combining it with training produced the best performance consistently.  As shown in Table~\ref{tab:mt-overview} tasks that derived from code properties (usage (U) and hierarchy (H)) did not benefit from training on ranking (R), forum to class (F) or linked posts (L) tasks on BERTOverflow, which suggests that tasks derived from code properties require different features than those emphasized by RFL tasks.  Tasks derived from code properties (HU) however helped RFL tasks, suggesting the importance of having a diversity of tasks for tuning.  We were expecting and found class hierarchy training to help the usage task, since code that is closely-related in the type hierarchy tends to have similar usage due to the nature of classes; this was confirmed by our findings.  We also expected usage analysis to help class hierarchy training, because we expected parameter types of methods to relate to the class hierarchy; this did not happen, perhaps due to the dynamically-typed nature of Python, where distinct types can share method names.  We expect this to be different in typed languages such as Java, and we plan to investigate it in our future work.


\begin{table}[]
\centering
\resizebox{1\columnwidth}{!}{%
\begin{tabular}{lrrrrr}
\toprule
Model       &           R  & F & L & H & U                                       \\
\midrule    

RFLHU-BERTOverflow & \bf 0.69/0.89 & 46.49 & \bf 198.10 & 0.15 & 0.27\\
RFLHU-CodeBERT & 0.68/0.88  & \bf 53.98 & 148.72 & 0.12 &  0.38\\
RFLH-BERTOverflow &0.68/0.89  & 47.56 &  188.73 & 0.17 & 0.14 \\
RFLH-CodeBERT & 0.67/0.88 & 49.04 & 141.97 & 0.10 & 0.14\\
RFL-BERTOverflow & 0.68/0.88 & 48.81 & 197.63 & 0.13 & 0.25\\
RFL-CodeBERT & 0.68/0.89 & 53.29 & 144.17 & 0.09 & 0.26\\
HU-BERTOverflow & 0.59/0.84 & 15.89 & 68.43 & \bf 0.29 & \bf 0.61 \\
HU-CodeBERT & 0.61/0.85 & 12.95 & 45.50 & 0.05 & 0.41\\

\bottomrule
\end{tabular}}
\caption{Answer Ranking (R) numbers are MRR/NDCG, Hierarchy (H) and Usage (U) tasks are correlation where as Linked Posts (L) and Forum to Doscstrings (F) are T-statistic. }
\label{tab:mt-overview}
\end{table}

\section{Conclusions}
In this paper, we presented BLANCA, a set of benchmarks to help further research in code understanding from textual manuals and posts about code.  We used BLANCA tasks to show that one can build better language models for understanding code artifacts. We also used multi-task training to demonstrate better representations of classes and functions from these models.  We hope these will be useful in enriching code representations with their textual semantics embedding in natural language artifacts of code.



\normalsize
\bibliography{anthology,custom}


\end{document}